\documentclass[letterpaper, 10pt, conference]{ieeeconf}
\IEEEoverridecommandlockouts
\IfFileExists{latexml.sty}{\usepackage{latexml}}{\newif\iflatexml}
\iflatexml\newcommand{\subfigref}[2]{\ref{#1}#2}\else\newcommand{\subfigref}[2]{\ref{#1#2}}\fi

\usepackage{amsmath,amsfonts}
\usepackage{algorithmic}
\usepackage{algorithm}
\usepackage{array}
\usepackage{tabularx}
\usepackage{textcomp}
\usepackage{xcolor}
\usepackage{stfloats}
\usepackage{cuted}
\usepackage{url}
\usepackage{verbatim}
\usepackage{graphicx}
\usepackage[caption=false,font=footnotesize]{subfig}
\usepackage{cite}
\usepackage{threeparttable}

\usepackage{booktabs}
\usepackage{siunitx}
\usepackage{multirow}
\usepackage{bm}
\usepackage{amssymb,stmaryrd}
\usepackage{gensymb}
\usepackage{leftindex}
\usepackage{times}
\usepackage{multicol}
\usepackage[
  bookmarks=true,
  hidelinks,
  pdftitle={ToF-Wheel Odometry: Navigating Degeneration of Localization with Limited Sensing Capabilities},
  pdfauthor={Xu Yang, Wenhao Li, Zhengyu Wei, Zening Li, Longxiang Huang, Bo Wang},
  pdfsubject={Accepted manuscript, IEEE Robotics and Automation Letters, DOI 10.1109/LRA.2026.3726370},
  pdfkeywords={Wheeled Robots, Localization, SLAM, Odometry}
]{hyperref}

\title{ToF-Wheel Odometry: Navigating Degeneration of Localization with Limited Sensing Capabilities%
\thanks{\iflatexml The authors are with Deptrum Ltd. \fi This is the authors' accepted manuscript of the article published in \emph{IEEE Robotics and Automation Letters}, vol.~11, no.~10, Oct.~2026, DOI: \href{https://doi.org/10.1109/LRA.2026.3726370}{10.1109/LRA.2026.3726370}.}%
\thanks{\textcopyright~2026 IEEE. Personal use of this material is permitted. Permission from IEEE must be obtained for all other uses, in any current or future media, including reprinting/republishing this material for advertising or promotional purposes, creating new collective works, for resale or redistribution to servers or lists, or reuse of any copyrighted component of this work in other works.}}

\begin{document}
\iflatexml
\author{Xu Yang, Wenhao Li, Zhengyu Wei, Zening Li, Longxiang Huang, and Bo Wang}
\maketitle
\else
\author{Xu Yang, Wenhao Li, Zhengyu Wei, Zening Li, Longxiang Huang, and Bo Wang\\
Deptrum Ltd.}
\maketitle
\fi
\bstctlcite{IEEEexample:BSTcontrol}

\begin{abstract}
Wheel odometry provides an efficient motion prior for ground robots but is prone to transient violations such as wheel slip or terrain-induced tilt.
In filtering-based odometry, these violations frequently trigger divergence when they coincide with constrained geometric observability (e.g., featureless corridors).
We address this challenge by proposing ToF--Wheel Odometry (TWO), a tightly-coupled filtering framework that integrates wheel kinematics with geometric constraints from Time-of-Flight (ToF) cameras.
The primary contribution of this work is a degeneracy-aware, directional covariance-shaping strategy that replaces heuristic online covariance tuning with geometry-conditioned inflation.
The core idea is to infer process noise from geometric corrections with a strict reliability check.
We decompose the posterior-to-prior correction into a wheel-explainable subspace and an orthogonal remainder.
The process covariance is anisotropically adjusted according to the observability of the ToF update: if the geometry is degenerate in non-kinematic directions, the remainder is suppressed as sensor noise; if the geometry is distinct, it is admitted as a valid kinematic violation.
Additionally, we introduce an intra-frame feedback loop that retrospectively repairs the process covariance within the current frame to reduce structural lag during bursty disturbances.
Extensive experiments on challenging scenarios demonstrate that TWO improves trajectory accuracy and robustness compared to static and adaptive Gaussian filtering, robustly navigating the trade-off between kinematic reliability and limited sensing capabilities.
\end{abstract}

\iflatexml
\else
\fi

\section{Introduction}\label{Introduction}

\begin{figure}[!t]
\centering
\iflatexml
\includegraphics[width=\linewidth]{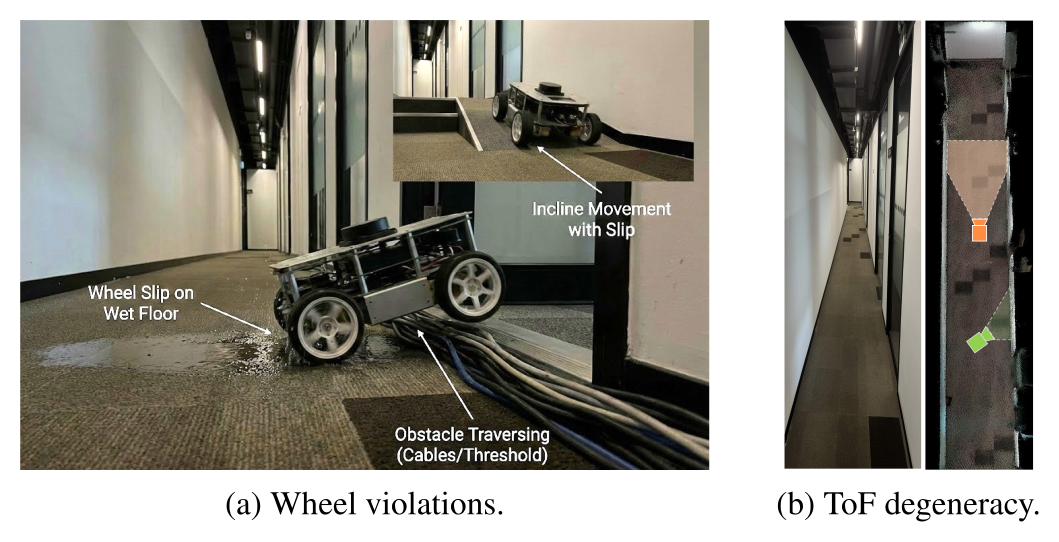}
\else
\subfloat[Wheel violations.]{%
\includegraphics[width=166pt]{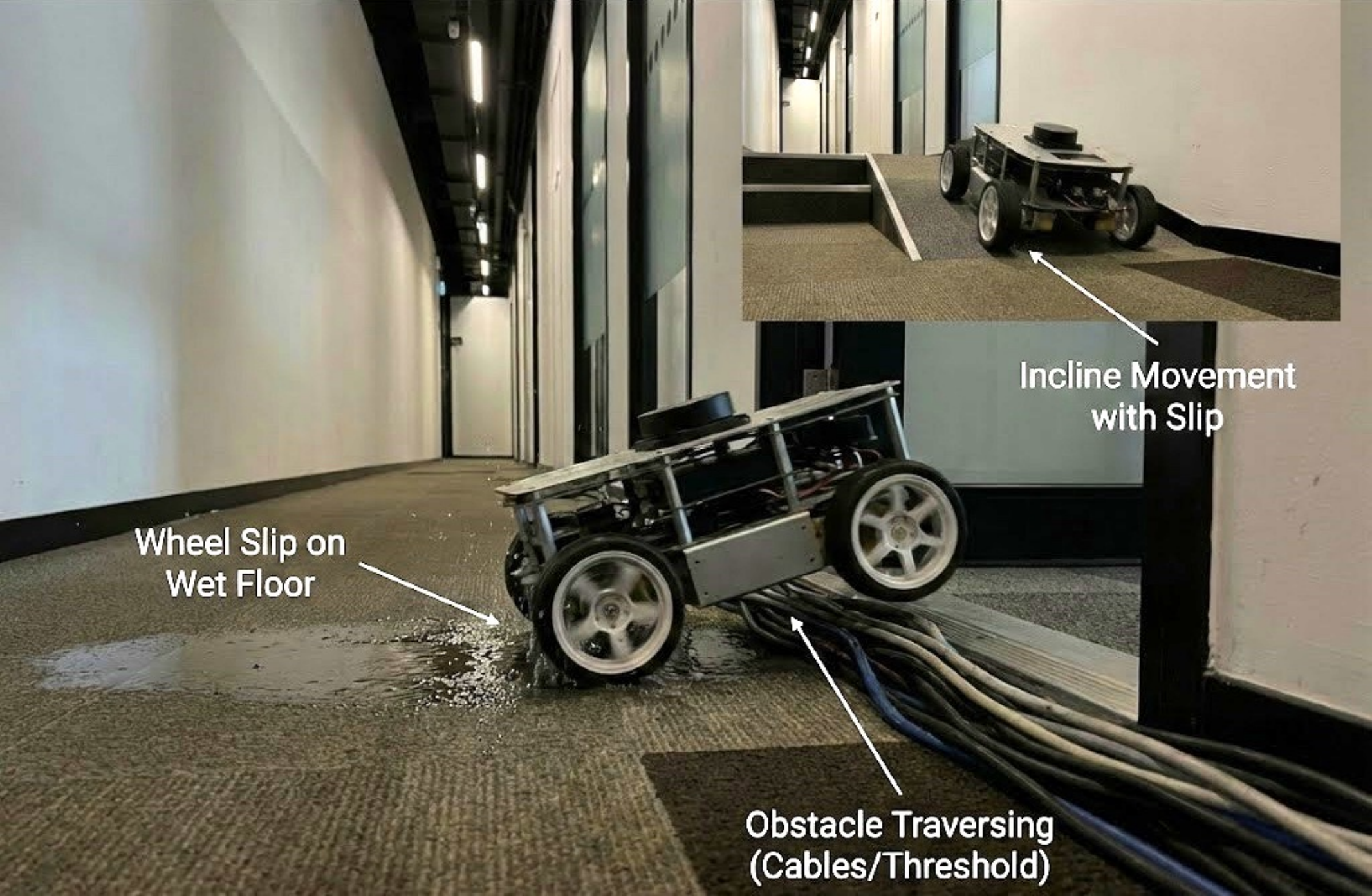}\label{fig:fig1a}}
\hfill
\subfloat[ToF degeneracy.]{%
\hspace{8pt}\includegraphics[width=60pt]{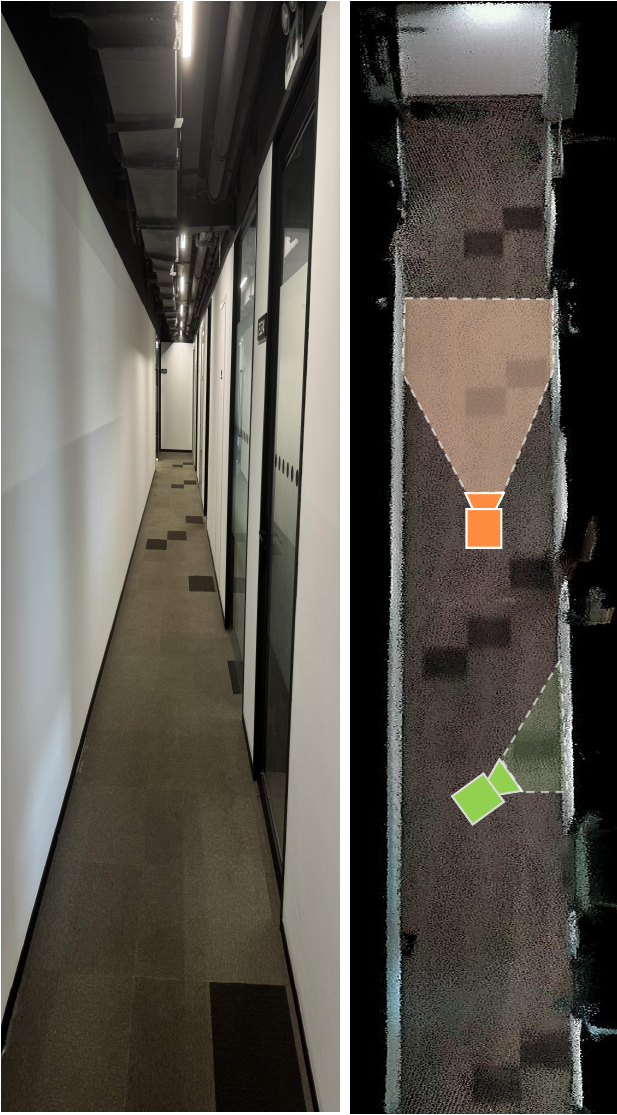}\hspace{8pt}\label{fig:fig1b}}
\fi
\caption{Motivating challenges in ToF--wheel odometry:
(a) wheel-odometry violations during terrain transitions, and (b) a top-down corridor point cloud with the ToF camera's narrow, short-range forward-facing field of view, shown at two robot positions.}
\label{fig:fig1}
\end{figure}

Ground robots commonly utilize wheel encoders as an efficient motion prior, favored for their low latency, simplicity, and robust short-term consistency~\cite{6792214,ROSE}.
Nevertheless, wheel motion models are inherently susceptible to validity violations such as slipping and tilting when traversing uneven terrain, slopes, or surface transitions (Fig.~\subfigref{fig:fig1}{a}).
In these scenarios, the foundational assumptions of planar motion and locally consistent attitude---especially roll and pitch---are compromised, resulting in transient yet significant discrepancies between the predicted and actual trajectories.

To improve robustness amid motion uncertainty, wheel encoders are routinely fused with exteroceptive geometric observations within a filtering framework~\cite{VINSwheels,LIWO,RGBDWheelSE2Constraints2021}.
On consumer-grade robots, Time-of-Flight (ToF) cameras are a common choice thanks to their compact size, low power consumption, and cost-effectiveness~\cite{ToFVO2017}.
However, ToF cameras have limited sensing range and only front-facing FoV, which restrict their geometric observability (Fig.~\subfigref{fig:fig1}{b}).
Consequently, localization degeneration often emerges from the interplay between limitations of these two sensors, rather than the failure of a single modality.

The inherent limitations of ToF cameras and wheel encoders pose a distinct challenge from those encountered in LiDAR--Inertial Odometry (LIO).
LIO typically benefits from robust geometric observation provided by wide-FoV, long-range LiDARs, complemented by 6-DoF motion priors from an IMU.
In contrast, TWO operates with a locally confined geometric observer (ToF) and a wheel-based motion prior that is prone to transient violations.

To mitigate localization degeneration under limited sensing, filtering-based odometries typically employ probabilistic motion models to regulate prediction uncertainty~\cite{FAST-LIO2,I2EKF}.
Most existing approaches assume Gaussian process noise with static~\cite{FAST-LIO2,VOXELMAP} or adaptive covariance~\cite{xie2025akf,I2EKF}.
Although adaptive Kalman filtering enhances robustness, many implementations rely on variance-only tuning (e.g., scalar or diagonal inflation) within a single Gaussian process model.
This assumption, however, proves inadequate for ground robots traversing terrain transitions, where wheel-induced roll and pitch errors manifest as transient, bursty disturbances rather than persistent noise.
Such non-stationary behavior defies effective management via uniform variance inflation: globally increasing uncertainty degrades nominal estimation confidence, whereas maintaining tight covariance results in dangerously overconfident predictions during critical transitions.
Furthermore, measurement-side robustification alone cannot resolve this issue, as the underlying uncertainty stems from violations of the motion model itself.

In this paper, we propose ToF--Wheel Odometry (TWO), a tightly-coupled filtering framework designed to explicitly mitigate motion-side uncertainty in wheel-driven ground robots subject to constrained geometric observability.
For TWO, robust process adaptation is not merely about how much uncertainty to add---it is also about where to place it, whether the geometry supports it, and when it must take effect.
TWO addresses this with a \emph{directional} covariance adaptation strategy, gated by observability scores that guard against ill-conditioned geometry.

The main contributions of this paper include:
\begin{enumerate}
\item \textbf{ToF--Wheel Odometry}: We present a tightly-coupled odometry framework that integrates wheel kinematics with ToF geometric constraints.
TWO targets robustness under kinematic violations and observation limitations.
\item \textbf{Degeneracy-aware Observability}: We derive geometric observability scores as bounded scalar approximations of the corresponding information-gain operators.
These scores function as reliability gates on covariance adaptation, suppressing correction components that lack local geometric support.
\item \textbf{Directional Covariance Shaping}: We introduce an orthogonal subspace decomposition strategy to adapt process covariance in kinematic and non-kinematic subspaces.
This enables anisotropic inflation of the process uncertainty according to observability scores.
\item \textbf{Intra-frame Feedback}: We introduce an intra-frame feedback loop to update process uncertainty retrospectively within the same frame.
This mitigates the structural lag inherent in standard Kalman filters, helping prevent divergence due to bursty disturbances.
\end{enumerate}

\begin{figure*}[t]
\centering
\includegraphics[width=\textwidth]{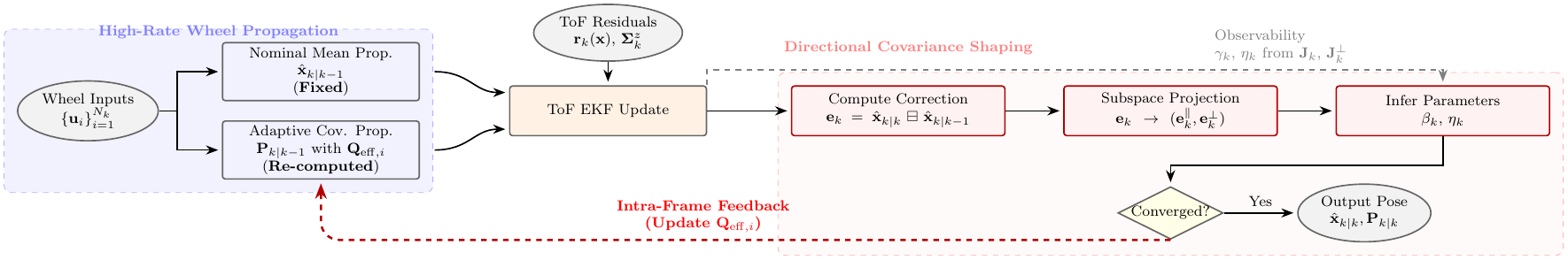}
\caption{Overview of the TWO pipeline. High-rate wheel propagation provides a motion prior, ToF EKF updates supply geometric corrections, and an intra-frame feedback loop performs degeneracy-aware directional covariance shaping.}
\label{fig:two_pipeline}
\end{figure*}

\section{Related Work}

\subsection{Wheel--Geometric Odometry}

In visual--inertial odometry, wheel constraints have been incorporated as auxiliary motion measurements or as preintegrated factors~\cite{VINSwheels}.
Recent work has also explored wheel-centric inertial sensing and learning-based priors to improve robustness under ground contact constraints, such as wheel-mounted IMU SLAM~\cite{Wheel-SLAM} and wheel--inertial neural odometry with ground manifold constraints~\cite{WING}.
In parallel, Potokar \emph{et al.}~\cite{ROSE} develop an on-manifold $\mathrm{SE}(3)$ wheel preintegration factor for off-road vehicles that estimates slip and wheel intrinsics online, with robust regularization to prevent slip from acting as an arbitrary slack variable.

In practice, the efficacy of such wheel factors remains contingent upon meticulous hyperparameter tuning to balance nominal accuracy and slip robustness.
TWO targets this gap by adapting \emph{process-side} uncertainty from ToF corrections and shaping covariance to physically meaningful subspaces.

In LiDAR--inertial odometry, wheel constraints are also integrated as an auxiliary motion source, e.g., LIWO~\cite{LIWO} and LIO~\cite{LIO2024}.
While these implementations underscore the practical utility of wheel priors, they typically rely on exteroceptive sensors (LiDARs) that are significantly more robust than the ToF cameras considered in this work.

Depth cameras provide local geometric constraints analogous to LiDAR registration but with limited range and field of view.
To mitigate these limitations, tightly coupled wheel--depth systems have been proposed to jointly exploit planar motion constraints with depth geometry (e.g., SE(2) plane constraints for RGB-D and wheel fusion~\cite{RGBDWheelSE2Constraints2021}).
Furthermore, ToF cameras have been utilized as auxiliary depth observers within visual odometry pipelines~\cite{ToFVO2017}.
Collectively, these works motivate the adoption of ToF as a compact geometric observer while simultaneously highlighting the need for motion models that remain robust under terrain-induced kinematic violations.

\subsection{Adaptive Kalman Filtering}

Adaptive Kalman filtering mitigates uncertainty misspecification by dynamically adjusting process or measurement covariances~\cite{mehra1970akf,myers1976adaptive,shumway1982em}.
Bayesian formulations provide a principled route to infer unknown noise parameters by assigning priors to covariances or scale variables and sequentially updating their posteriors~\cite{VB2017HUANG,BBVIAKF2020}.
In robotics, these methodologies are often implemented as adaptive error-state filters or iterated EKF variants to improve stability~\cite{I2EKF}.

Many practical adaptive filtering methods retain a single-Gaussian process model, focusing predominantly on variance modulation to handle non-stationarity.
As discussed in Sec.~\ref{Introduction}, this is a poor match for the bursty, high-amplitude violations that terrain transitions and traction loss produce on ground robots: globally inflating covariance to accommodate these events compromises nominal accuracy, while keeping it small causes overconfidence during transitions.

\subsection{Robust Estimation and Process-side Robustness}
Standard robust estimation techniques typically focus on measurement-side outlier rejection to mitigate the influence of non-Gaussian noise~\cite{KhosoussiK-RSS-25}.
However, ground robots frequently encounter process-side non-stationary disturbances—such as wheel slip or terrain-induced tilt—that manifest as transient, high-amplitude bursts.
Recent advancements have investigated robust process-noise modeling for Lie-group estimation~\cite{IKFHeavyTail2021,RobustKFHeavyTail2021}.
In contrast to these works, the proposed framework does not introduce a novel non-Gaussian process.
Instead, TWO employs a degeneracy-aware, directionally structured covariance shaping strategy.

\subsection{Geometric Degeneracy and Localizability}
Geometric constraints can become ill-conditioned in environments where the observation fails to provide sufficient excitation along specific degrees of freedom.
Recent work in registration and calibration has therefore emphasized localizability and observability analysis to identify unreliable estimation directions~\cite{X-ICP,OBS-AWARE}.
In TWO, we follow this line of reasoning but in a different way: instead of modifying the measurement model, we compute observability scores that function as continuous quality knobs to gate remainder admission and to damp weight updates under poor observability.

\section{System Overview}\label{sec:system_overview}
We consider a ground robot equipped with differential-drive wheel encoders and a forward-facing ToF camera.
To robustly handle terrain-induced motion violations, TWO extends the standard EKF with degeneracy-aware directional covariance shaping and an intra-frame feedback loop.
Figure~\ref{fig:two_pipeline} summarizes the four-stage pipeline: wheel propagation, ToF update with observability analysis, directional covariance shaping, and intra-frame feedback.

\textbf{Wheel propagation.}
Between successive ToF frames, the pose is propagated using a kinematic model driven by wheel measurements.
In particular, the process covariance is propagated with
\begin{equation}
\mathbf{P}_{i+1}=\mathbf{F}_i\mathbf{P}_{i}\mathbf{F}_i^\top+\mathbf{Q}_{\mathrm{eff},i},
\label{eq:propagate_cov}
\end{equation}
Here $\mathbf{F}_i$ is the state-transition Jacobian of the standard differential-drive model, and $\mathbf{Q}_{\mathrm{eff},i}$ is the corresponding process-covariance increment.
It is the sum of three terms: the mapped wheel-encoder covariance $\mathbf{Q}_{w,i}$, a constant baseline disturbance covariance $\mathbf{Q}_0$, and an inflated covariance inferred from the ToF update $\mathbf{Q}_{\mathrm{infl},i}$:
\begin{equation}
\mathbf{Q}_{\mathrm{eff},i} \triangleq \mathbf{Q}_{w,i} + \mathbf{Q}_0 + \mathbf{Q}_{\mathrm{infl},i}.
\label{eq:qeff_components}
\end{equation}

\textbf{ToF update.}
We utilize probabilistic point-to-plane constraints~\cite{VOXELMAP,C3P} to form a residual vector $\mathbf{r}_k(\mathbf{x})$ with covariance $\bm{\Sigma}^z_k$.
We denote its stacked Jacobian at the current estimate by $\mathbf H_k$.
The MAP estimate is obtained by minimizing the joint cost of prior and measurement residuals:
\begin{equation}
\begin{split}
\hat{\mathbf{x}}_{k|k}
&=\arg\min_{\mathbf{x}\in\mathrm{SE}(3)}\\
&\quad\Big(\left\|\mathbf{x}\boxminus\hat{\mathbf{x}}_{k|k-1}\right\|^2_{\mathbf{P}_{k|k-1}^{-1}}
+ \left\|\mathbf{r}_k(\mathbf{x})\right\|^2_{(\bm{\Sigma}^z_k)^{-1}}\Big).
\end{split}
\label{eq:tof_map_cost}
\end{equation}

The correction between the prior and the posterior is
\begin{equation}
\mathbf{e}_k = \hat{\mathbf{x}}_{k|k} \boxminus \hat{\mathbf{x}}_{k|k-1}.
\label{eq:posterior_correction}
\end{equation}

\textbf{Directional covariance shaping.}
Once a posterior correction $\mathbf{e}_k$ is computed, TWO does not simply apply it as a one-time fix.
Instead, it analyzes why the posterior differs from the prior and shapes the process covariance to anticipate similar errors.
The goal is to expand the prior uncertainty along the specific directions of motion-model violation, but only to the extent supported by valid geometric measurements.

The correction vector $\mathbf{e}_k$ is decomposed to distinguish between errors the wheels can explain $\mathbf{e}_k^{\parallel}$ and errors they cannot $\mathbf{e}_k^{\perp}$.
We write the correction in the body frame as $\mathbf{e}_k=[\phi_x,\phi_y,\phi_z,\rho_x,\rho_y,\rho_z]^\top$.
For a differential-drive platform with track width $l$, the wheel-explainable basis is
\begin{equation}
\mathbf{B}_{\mathrm{dd}}=
\begin{bmatrix}
0&0&-1/l&1/2&0&0\\
0&0& 1/l&1/2&0&0
\end{bmatrix}^{\!\top}.
\label{eq:dd_basis}
\end{equation}
We project $\mathbf{e}_k$ onto a wheel-explainable subspace and an orthogonal remainder:
\begin{equation}
\begin{aligned}
\mathbf c_k
&=(\mathbf B_{\mathrm{dd}}^\top\mathbf B_{\mathrm{dd}})^{-1}
\mathbf B_{\mathrm{dd}}^\top\mathbf e_k,\\
\mathbf e_k^{\parallel}&=\mathbf B_{\mathrm{dd}}\mathbf c_k,\qquad
\mathbf{e}_k^{\perp}=\mathbf{e}_k-\mathbf{e}_k^{\parallel}.
\end{aligned}
\label{eq:correction_decomposition}
\end{equation}
Since $\operatorname{Col}(\mathbf B_{\mathrm{dd}})$ spans the yaw and forward directions, Eq.~\eqref{eq:correction_decomposition} simply retains these two components.
\begin{itemize}
    \item The wheel-explainable component $\mathbf{e}_k^{\parallel}$ is the projection onto the manifold of valid kinematic motions (forward, backward, yaw rotation).
    \item The orthogonal remainder $\mathbf{e}^{\perp}_k$ is the component perpendicular to that manifold (e.g., lateral sliding, vertical motion, or roll/pitch tilt), representing motion explicitly forbidden by the standard kinematic model.
\end{itemize}
This separation allows TWO to treat kinematic violations (e.g., wheel slip affecting forward/yaw motion) differently from extrinsic disturbances (e.g., terrain-induced tilt).

Based on this decomposition, TWO adjusts the magnitude and structure of the process covariance $\mathbf{Q}_{\mathrm{infl},i}$ according to geometric observability.

\textbf{Intra-frame feedback.}
In a conventional filter, the estimation for frame $k$ concludes after the update.
However, a large correction $\mathbf{e}_k$ implies that the prior $\hat{\mathbf{x}}_{k|k-1}$ and its covariance $\mathbf{P}_{k|k-1}$ were physically inconsistent with reality.

TWO addresses this via an intra-frame feedback loop.
Instead of moving immediately to frame $k+1$, the system repairs the prior by re-propagating the inferred $\mathbf{Q}_{\mathrm{infl},i}$.
This mechanism aligns the filter's uncertainty with the observation before future integration.

\section{Degeneracy-aware Observability Analysis}
\label{sec:observability_metrics}

ToF cameras have limited range and field-of-view, which can lead to degenerate observations.
Adapting the covariance based on ill-conditioned geometry (e.g., long corridors) can cause the filter to drift or diverge.
To adapt the process covariance robustly, we estimate two observability scores for each ToF frame.

\subsection{Geometric Observability}
\label{subsec:geometric_observability}

We quantify geometric observability by analyzing the information gain provided by the ToF measurement, given by the measurement information matrix
\begin{equation}
\mathbf{J}_k\triangleq \mathbf{H}_k^\top(\bm{\Sigma}^z_k)^{-1}\mathbf{H}_k.
\label{eq:measurement_information}
\end{equation}
where the columns of $\mathbf H_k$ are scaled by fixed rotation--translation weights so that rotational and translational directions contribute on a comparable numerical scale.
We define the corresponding \emph{gain operator} as
\begin{equation}
\mathbf{K}_k \triangleq \mathbf{J}_k (\mathbf{I} + \mathbf{J}_k)^{-1}.
\label{eq:posterior_gain_operator}
\end{equation}
Let $\mathbf{J}_k = \mathbf{U} \text{diag}(\lambda_{k,j}) \mathbf{U}^{\top}$ be its eigendecomposition.
The information-gain operator can be expressed spectrally as
\begin{equation}
\mathbf{K}_k = \mathbf{U} \text{diag}\left( \frac{\lambda_{k,j}}{\lambda_{k,j} + 1} \right) \mathbf{U}^{\top}.
\label{eq:gain_operator_spectrum}
\end{equation}
Each gain $\frac{\lambda_{k,j}}{\lambda_{k,j} + 1}$ approaches one when a direction is strongly measured and well constrained by the ToF update, and approaches zero when it is weakly measured and left largely unconstrained.

While the full matrix $\mathbf{K}_k$ captures anisotropic observability, using a full matrix to gate covariance adaptation is computationally complex and potentially unstable.
Instead, we seek a scalar score $\gamma_k$ that serves as the optimal isotropic approximation of $\mathbf{K}_k$.
We formulate this as a least-squares minimization in the Frobenius norm:
\begin{equation}
\gamma_k = \underset{\gamma}{\arg\min} \| \gamma \mathbf{I} - \mathbf{K}_k \|_F^2.
\label{eq:gamma_ls_projection}
\end{equation}
The closed-form solution to this optimization is the normalized trace of the gain operator.
Thus, we define the geometric observability score $\gamma_k$ as

\begin{equation}
\gamma_k \triangleq \frac{1}{\dim(\mathbf{J}_k)} \text{tr}(\mathbf{K}_k) = \frac{1}{\dim(\mathbf{J}_k)} \sum_{j=1}^{\dim(\mathbf{J}_k)} \frac{\lambda_{k,j}}{\lambda_{k,j} + 1} \in [0, 1].
\label{eq:gamma_def}
\end{equation}

\begin{itemize}
    \item $\gamma_k \to 1$: The geometry is spatially diverse (e.g., a cluttered room), effectively ``locking'' all six DoFs.
    We fully trust the ToF update to drive the process adaptation.
    \item $\gamma_k \to 0$: The geometry provides little information across most directions.
    A long corridor, for instance, weakens the longitudinal direction specifically. This moderately lowers $\gamma_k$, and the reduction grows as more directions become weakly observed.
    In this case, we damp the inflation magnitude to prevent the filter from overfitting the noise along the unconstrained direction.
\end{itemize}
\subsection{Remainder-Subspace Observability}
\label{subsec:remainder_subspace_obs}

To evaluate the reliability of measurements in directions orthogonal to the wheel kinematics, we restrict the same information construction to the orthogonal complement spanned by $\mathbf{U}_{\perp}$.
We use a fixed orthonormal basis $\mathbf{U}_{\perp}\in\mathbb{R}^{6\times4}$ for the remaining four directions.
We define the projected measurement information matrix as $\mathbf{J}_{\perp,k} \triangleq \mathbf{U}_{\perp}^{\top}\mathbf{J}_k\mathbf{U}_{\perp}$ and the corresponding information-gain operator as
\begin{equation}
\mathbf{K}_{\perp,k} \triangleq \mathbf{J}_{\perp,k}(\mathbf{I} + \mathbf{J}_{\perp,k})^{-1}.
\label{eq:subspace_gain_operator}
\end{equation}

Applying the same Frobenius-norm definition, we obtain the remainder gate $\eta_k$:
\begin{equation}
\begin{split}
\eta_k
&\triangleq \frac{1}{\dim(\mathbf{J}_{\perp,k})}\text{tr}(\mathbf{K}_{\perp,k}) \\
&= \frac{1}{\dim(\mathbf{J}_{\perp,k})}
\sum_{j=1}^{\dim(\mathbf{J}_{\perp,k})}
\frac{\lambda_{k,j}^{\perp}}{\lambda_{k,j}^{\perp} + 1} \in [0, 1].
\end{split}
\label{eq:eta_def}
\end{equation}
where $\lambda_{k,j}^{\perp}$ are the eigenvalues of $\mathbf{J}_{\perp,k}$.
This average remainder-subspace information gain acts as a reliability gate:

\begin{itemize}
    \item $\eta_k \to 1$: The geometry provides strong constraints in non-kinematic directions.
    \item $\eta_k \to 0$: The geometry is degenerate in non-kinematic directions.
\end{itemize}

\section{Directional Covariance Shaping}
\label{sec:dir_shaping}

We derive two scalar parameters to adjust the magnitude and structure of the process covariance.

\subsection{Kinematic Magnitude Adaptation}
We analyze the two wheel-explainable components $\mathbf c_k$ defined in Eq.~\eqref{eq:correction_decomposition}.
Physically, a large $\mathbf c_k$ indicates that the observed motion deviates significantly from what the wheel encoder predicts.
We model this mismatch by applying a scale factor $(1+\beta_k)$ to inflate the covariance along the kinematic manifold:
\begin{equation}
\mathbf c_k\sim\mathcal N\!\left(
\mathbf0,(1+\beta_k)\bm\Sigma^{\parallel}_{u,k}
\right).
\label{eq:kinematic_correction_likelihood}
\end{equation}
where $\bm\Sigma^{\parallel}_{u,k}$ is the prior-to-posterior covariance reduction $\mathbf P_{k|k-1}-\mathbf P_{k|k}$, projected onto the same wheel-explainable subspace as $\mathbf c_k$. Intuitively, it captures how much of this frame's correction the wheel model already explains.
We estimate $\beta_k$ by minimizing the negative log-likelihood of observing $\mathbf c_k$:
\begin{equation}
\mathcal{L}_{\beta}(\beta_k)
= \frac{1}{2}\frac{\psi^{\parallel}_k}{1+\beta_k}
+\frac{d_{\parallel}}{2}\log(1+\beta_k),
\quad \beta_k\ge 0,
\label{eq:beta_type2_main}
\end{equation}
where $\psi^{\parallel}_k = \mathbf{c}_k^\top(\bm{\Sigma}^{\parallel}_{u,k})^\dagger\mathbf{c}_k$ and $d_{\parallel}=\operatorname{rank}(\bm{\Sigma}^{\parallel}_{u,k})$ are its Mahalanobis energy and effective rank, respectively, and $(\cdot)^\dagger$ denotes the Moore--Penrose pseudoinverse.

The resulting closed-form minimizer is
\begin{equation}
\beta_k^\star=\max(0,\frac{\psi^{\parallel}_k}{d_{\parallel}}-1).
\label{eq:beta_star}
\end{equation}
The truncation $\max(0,\cdot)$ enforces that we only \emph{inflate} uncertainty but never deflate it based on finite-sample fluctuations.

To prevent spurious inflation in degenerate environments where the correction itself may be unreliable, we dampen the update with the geometric observability score $\gamma_k$:
\begin{equation}
\beta_k \leftarrow (1-\gamma_k)\beta_{k-1} + \gamma_k\max(0, \frac{\psi^{\parallel}_k}{d_{\parallel}} - 1).
\label{eq:beta_soft_update_unified}
\end{equation}
This ensures that we only expand process uncertainty when the geometric observability is sufficient to validate the kinematic violation.

\subsection{Remainder Subspace Gating}

The orthogonal remainder $\mathbf{e}_k^{\perp}$ contains posterior-correction components outside the differential-drive subspace, such as lateral, vertical, or roll/pitch components.
Its magnitude alone, however, does not indicate how reliably these components are supported by the current ToF observations.
The remainder subspace's reliability depends on the local geometry: when it is weakly constrained, sensor noise can be amplified into a large $\mathbf{e}_k^{\perp}$.
We therefore use $\eta_k$ (Eq.~\eqref{eq:eta_def}) as a continuous aggregate geometric-support weight:
\begin{equation}
\tilde{\mathbf{e}}_k^{\perp} \triangleq \eta_k \mathbf{e}_k^{\perp}.
\label{eq:gated_remainder}
\end{equation}

This gating strategy effectively navigates the trade-off between model rigidity and measurement trust:
\begin{itemize}
    \item Strong geometric support ($\eta_k \to 1$): The geometry provides sufficient constraints to distinguish physical motion from noise.
    The remainder is treated as a valid kinematic violation and fully admitted.
    \item Weak geometric support ($\eta_k \to 0$): The observation is unreliable in the non-kinematic subspace.
    The remainder is suppressed as sensor noise to prevent incorporating phantom corrections into the process model.
\end{itemize}

\subsection{Directional Inflation}
Finally, we reconstruct a reliability-filtered correction by recombining the kinematic component with the gated remainder
\begin{equation}
\tilde{\mathbf{e}}_k \triangleq \mathbf{e}^{\parallel}_k + \tilde{\mathbf{e}}^{\perp}_k
\label{eq:tilde_e_map}
\end{equation}
and normalize it as
\begin{equation}
\mathbf{s}_k \triangleq
\begin{cases}
\tilde{\mathbf{e}}_k/\|\tilde{\mathbf{e}}_k\|_{\mathbf{Q}_{\mathrm{ref}}^{-1}}, & \tilde{\mathbf{e}}_k\neq \mathbf{0},\\
\mathbf{0}, & \text{otherwise},
\end{cases}
\label{eq:u_unified}
\end{equation}
where $\mathbf{Q}_{\mathrm{ref}}$ is a fixed reference covariance, so that $\mathbf{s}_k$ retains $\tilde{\mathbf{e}}_k$'s pose-tangent units.
The direction $\mathbf{s}_k$ determines how the additional process covariance is shaped during intra-frame re-propagation.
We distribute the frame-level inflation evenly over the $N_k$ cached wheel propagation steps, one for each wheel reading within frame $k$.
At step $i$, we transport the frame-end direction $\mathbf{s}_k$ back to the $i$-th wheel-propagation step, using the $\mathrm{SE}(3)$ adjoint of the relative pose $\Delta\hat{\mathbf{T}}_{i,k}$ between step $i$ and the frame end:
\begin{equation}
\mathbf{s}_{k,i}\triangleq\operatorname{Ad}_{\Delta\hat{\mathbf{T}}_{i,k}}\mathbf{s}_k,
\end{equation}
which gives
\begin{equation}
\mathbf{Q}_{\mathrm{infl},i} \triangleq \frac{\beta_k}{N_k}\,\mathbf{s}_{k,i}\mathbf{s}_{k,i}^\top.
\label{eq:qeff_unified}
\end{equation}
Only the wheel-explainable correction drives the raw update of $\beta_k$.
$\eta_k\mathbf e_k^\perp$ shapes $\mathbf{s}_k$ but cannot initiate inflation when $\beta_k=0$, because $\mathbf e_k^\perp$ alone may reflect geometric degeneracy or sensor error rather than genuine motion.
A purely orthogonal, wheel-silent disturbance may still be corrected by the ToF update but cannot independently trigger inflation.

\section{Intra-frame Feedback}

Standard Kalman filtering operates sequentially: the process noise is fixed a priori during propagation, and the state is updated a posteriori.
This creates a structural vulnerability.
When a sudden kinematic violation occurs (e.g., a wheel slip), the fixed prior covariance remains overconfident.
The filter underestimates the uncertainty exactly when the error is largest.

To resolve this, TWO implements an intra-frame feedback loop.
We treat the covariance adaptation not as a post-processing step, but as an integral part of the estimation convergence within the current frame.
Specifically, the updated process covariance $\mathbf{Q}_{\mathrm{eff},i}$ is fed back to the wheel propagation step and triggers a re-propagation and a re-update.
It ends when the posterior and prior converge or after a maximum number of iterations.
Thus, the extra latency is bounded by a small fixed number of EKF updates.
The process covariance is repaired within the same frame rather than after a multi-frame adaptive window.

During feedback, the propagated mean and local map remain fixed.
Each covariance replay starts from $\mathbf P_{k-1|k-1}$, and the map is updated only after a posterior is accepted.
The current-frame covariance is repaired by alternating a few times between re-solving Eq.~\eqref{eq:tof_map_cost} for the state estimate and updating the covariance parameters (the same state-then-hyperparameter alternation used in Empirical-Bayes adaptive filtering~\cite{mehra1970akf,myers1976adaptive,shumway1982em}). Each pass uses the current posterior-to-prior correction to revise those parameters.

\section{Experiments}
\label{sec:experiments}

As the focus of this work is to handle kinematic violations and geometric degeneracy, experiments are designed to expose these challenges.
The analysis therefore emphasizes extreme, bursty conditions rather than only aggregate average performance.

\subsection{Datasets}

We evaluate TWO on two datasets with adverse conditions:
\begin{itemize}
    \item \textit{Self-collected data}: Three long indoor/semi-structured sequences (Lobby, Office, Corridor) support cross-method comparison.
    Ablations use \emph{Slope} (incline), \emph{Forward Slip + Weak Geometry} (longitudinal slip with weak ToF geometry), \emph{Side Slip} (lateral off-manifold motion), and \emph{Lift/Pose Perturbation} (vertical/attitude disturbances).
    \item \textit{M3DGR public dataset}~\cite{zhang2025towards}: Public sequences involving kinematic challenges (Wheel Float, Sharp Turn).
\end{itemize}
The self-collected data were recorded on a differential-drive platform with a Raspberry Pi~4B for onboard computation.
For the self-collected sequences, ground-truth trajectories were obtained using an independent chassis-mounted Livox Mid-360 LiDAR--inertial system.
The Mid-360 measurements are used only for ground-truth generation and are not used by TWO.
All parameters are fixed across sequences.
\mbox{We use} $0.2\,\mathrm{rad\,s^{-1}}$ wheel-rate noise per wheel for $\mathbf{Q}_{w,i}$, $\mathbf{Q}_0=10^{-8}\operatorname{diag}(1,1,0,0,1,1)$, and $\mathbf{Q}_{\mathrm{ref}}=10^{-6}\mathbf{I}_6$ in the coordinate order above.
Eq.~\eqref{eq:measurement_information}'s rotation--translation weights use reference variances of $10^{-3}\,\mathrm{rad}^2$ and $2\times10^{-4}\,\mathrm m^2$.

\begin{figure*}[t]
\centering
\includegraphics[width=0.90\textwidth]{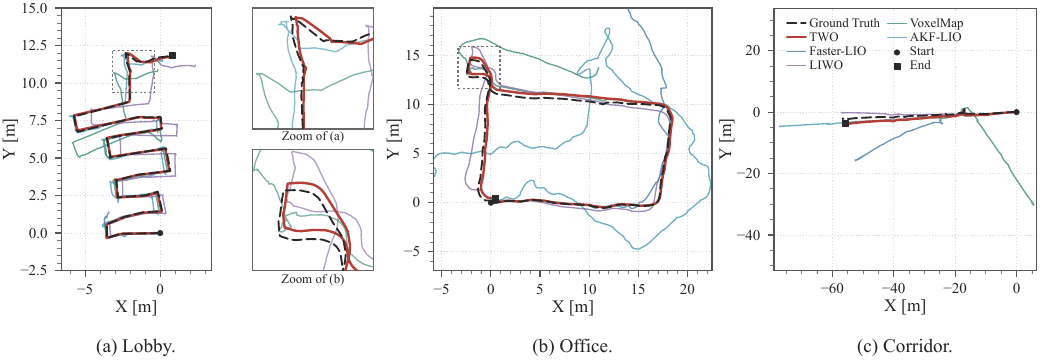}
\caption{Qualitative trajectory visualization on self-collected long sequences.
The two central close-ups enlarge the dashed regions in (a) and (b), respectively.}
\label{fig:selfcollected_traj}
\vspace{-0.5ex}
\end{figure*}

\begin{table}[t]
\centering
\caption{Self-collected/M3DGR ATE/RPE (best in bold).}
\scriptsize
\setlength{\tabcolsep}{2.1pt}
\renewcommand{\arraystretch}{1.15}
\begin{threeparttable}
\begin{tabular}{@{}lcccccc@{}}
\toprule
\multirow{2}{*}{\textbf{Method}}
& \multicolumn{5}{c}{\textbf{ATE / RPE@1m}}
& \multirow{2}{*}{\textbf{Avg.}}\\
\cmidrule(lr){2-6}
& Lobby & Office & Corridor & Wheel Float & Sharp Turn &\\
\midrule
\textbf{TWO}
& \textbf{0.072}/0.061 & \textbf{0.177}/\textbf{0.058} & \textbf{0.249}/\textbf{0.077} & 0.209/0.091 & 0.242/0.088 & \textbf{0.190}\\
VoxelMap
& 0.620/0.129 & 4.901/0.136 & 9.705/0.234 & 0.983/0.264 & 0.538/0.152 & 3.349\\
AKF-LIO
& 0.239/0.237 & 4.470/0.735 & 3.234/0.570 & \textbf{0.190}/\textbf{0.045} & \textbf{0.200}/\textbf{0.050} & 1.667\\
Faster-LIO
& -- & 11.232/0.973 & 2.290/0.558 & \textbf{0.190}/\textbf{0.045} & 0.201/\textbf{0.050} & 3.478\\
LIWO
& 0.270/\textbf{0.056} & 0.618/0.168 & 0.424/0.105 & 0.539/0.330 & 0.202/0.053 & 0.411\\
\bottomrule
\end{tabular}
\begin{tablenotes}[flushleft]\scriptsize
\item \textit{Note:} Drift failures omitted.
Avg.\ ATE is the arithmetic mean over available sequences; Faster-LIO's average excludes its omitted Lobby result.
\end{tablenotes}
\end{threeparttable}
\label{tbl:ate_rpe}
\end{table}

\subsection{Comparative Analysis}

We compare TWO with VoxelMap~\cite{VOXELMAP}, AKF-LIO~\cite{xie2025akf}, Faster-LIO~\cite{FASTER-LIO}, and LIWO~\cite{LIWO}.
TWO shares the VoxelMap/C3P-VoxelMap point-to-plane update backbone~\cite{VOXELMAP,C3P}, controlling for the geometric frontend in that comparison.

Fig.~\ref{fig:selfcollected_traj} compares the estimated and ground-truth trajectories in three indoor environments.
TWO preserves trajectory fidelity through sharp turns, large loops, and corridor degeneracy; the baselines show path distortion, rotational drift, or longitudinal overshoot.
Trajectories with severe drift are omitted for clarity.
Lobby and Office test recovery from turning and terrain transitions, while Corridor stresses weak longitudinal excitation.
In Lobby, TWO follows the zig-zagging ground truth through slip-affected turns.
In Office, it preserves the rectangular loop while VoxelMap and LIWO accumulate rotational drift.
In Corridor, it avoids baseline longitudinal overshoot by damping unreliable corrections under weak geometry.

Table~\ref{tbl:ate_rpe} shows that
TWO achieves the lowest average ATE across all five sequences. Notably, it does so without a gyroscope, while every baseline here has one.
On the self-collected sequences, TWO and the baselines see the same ToF point cloud, and only the motion prior differs, with the baselines propagating from the module's built-in IMU and TWO from the wheel encoders alone.
For M3DGR sequences, since the dataset provides no ToF camera at all, TWO's point-to-plane update instead runs on the same LiDAR stream the baselines use.

TWO leads VoxelMap by a wide margin on both sequences, resulting from the wheel prior VoxelMap lacks entirely.
It also beats LIWO on Wheel Float, where a floating wheel's readings turn misleading rather than merely noisy---precisely the failure directional gating is built to catch.
AKF-LIO and Faster-LIO propagate from a gyroscope, which directly measures the angular-rate signal TWO must instead infer from geometry.
This shows up unevenly: the baselines roughly halve TWO's RPE on Wheel Float ($0.045$ vs.\ $0.091$\,m) and cut it by over 40\% on Sharp Turn.
This gap reflects the gyroscope's direct advantage.
The ATE gap, however, is far narrower ($0.190$ vs.\ $0.209$\,m; $0.200$ vs.\ $0.242$\,m). Without a rotation-rate sensor, TWO's inferred covariance shaping still recovers comparable long-horizon accuracy.

To isolate the covariance-shaping mechanism from sensor configuration, Table~\ref{tbl:baseline_comparison} compares TWO with wheel-only, fixed-$\mathbf Q$ ToF--wheel, and scalar-$\mathbf Q$ ToF--wheel baselines.
The scalar-$\mathbf Q$ baseline rescales the nominal wheel-process covariance by matching its trace to the current correction energy, without directional shaping or feedback.
Scalar-$\mathbf Q$ underperforms even the static fixed-$\mathbf Q$ baseline on every metric (e.g., $0.295$ vs.\ $0.255$\,m ATE RMSE). Because it inflates uncertainty isotropically whenever a correction occurs, it injects uncertainty into directions the wheel prior already constrains well---degrading confidence exactly where the filter did not need help. This is the failure mode that directional shaping is designed to avoid.
This indicates that magnitude-only adaptation is not merely insufficient but can be actively counterproductive, motivating TWO's joint direction-and-magnitude adjustment.

\begin{table}[!t]
\centering
\begingroup
\caption{Comparison with wheel-only and fixed-/scalar-$\mathbf Q$ baselines.}
\label{tbl:baseline_comparison}
\scriptsize
\setlength{\tabcolsep}{4pt}
\renewcommand{\arraystretch}{1.08}
\begin{threeparttable}
\begin{tabular*}{\columnwidth}{@{\extracolsep{\fill}}lcccc@{}}
\toprule
\textbf{Method} & \textbf{ATE RMSE} & \textbf{ATE P95} & \textbf{RPE} & \textbf{Worst 3-s}\\
\midrule
Wheel-only & 0.628 & 1.170 & 0.118 & 1.179\\
ToF+Wheel fixed $\mathbf{Q}$ & 0.255 & 0.453 & 0.085 & 0.459\\
ToF+Wheel scalar $\mathbf{Q}$ & 0.295 & 0.464 & 0.087 & 0.474\\
\textbf{TWO} & \textbf{0.046} & \textbf{0.084} & \textbf{0.029} & \textbf{0.084}\\
\bottomrule
\end{tabular*}
\begin{tablenotes}[flushleft]\scriptsize
\item Accuracy entries are four-sequence means in meters.
\end{tablenotes}
\end{threeparttable}
\endgroup
\vspace{-1.8ex}
\end{table}

\begin{figure*}[t]
\centering
\includegraphics[width=0.67\textwidth]{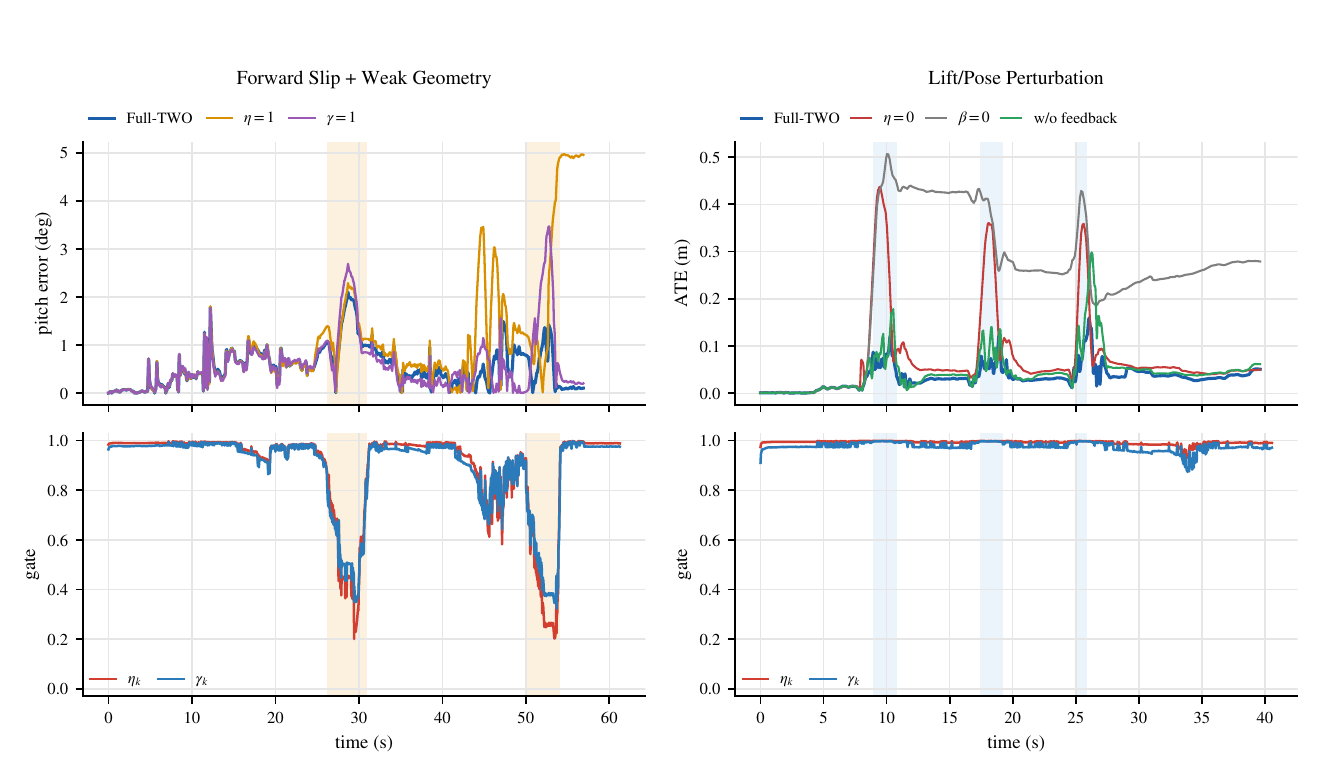}
\caption{Selective behavior of TWO under different physical events.
Left: under slip with degraded geometry, the pitch-error curve diagnoses attitude contamination; TWO lowers $\eta_k$ and $\gamma_k$ and avoids the larger attitude errors of fixed-gate variants.
Right: under observable Lift/Pose Perturbation events, TWO keeps the gates open and admits the geometrically supported off-manifold correction, while suppressing the remainder ($\eta=0$) or disabling inflation feedback ($\beta=0$) leads to larger trajectory errors.}
\label{fig:ablation_selective_behavior}
\end{figure*}

\subsection{Ablation Studies}

The ablations assess three design hypotheses concerning direction, magnitude, and timing: (C1) admit the non-kinematic remainder into the inflation direction only when supported by the local ToF geometry; (C2) damp the correction-driven inflation-magnitude update under weak geometry; and (C3) apply the resulting covariance within the same frame rather than at the next update.
We compare the complete system against five variants:
\begin{enumerate}
    \item \textit{TWO ($\eta=0$)}: Suppresses the orthogonal remainder entirely, enforcing rigid kinematic constraints.
    \item \textit{TWO ($\eta=1$)}: Admits the orthogonal remainder fully, bypassing the remainder gating mechanism regardless of ToF geometric constraints.
    \item \textit{TWO ($\gamma=1$)}: Disables degeneracy-aware damping, updating the inflation parameter $\beta$ directly based on the current correction.
    \item \textit{TWO ($\beta=0$)}: Fixes the inflation magnitude at zero, maintaining the nominal process covariance.
    \item \textit{TWO w/o feedback}: Exposes the one-frame delay inherent in standard filtering by utilizing the inferred covariance only after completing the current update.
\end{enumerate}

\textbf{Selective remainder admission (C1).}
Table~\ref{tbl:two_ablation_metrics} and Fig.~\ref{fig:ablation_selective_behavior} expose the two regimes that $\eta_k$ must distinguish.
During Forward Slip + Weak Geometry, the fixed $\eta=1$ variant admits unreliable non-kinematic corrections and yields a visibly higher attitude error.
During Side Slip and Lift/Pose Perturbation, by contrast, the off-manifold motion is geometrically supported. $\eta=1$ performs close to full TWO on these two sequences, reflecting proper gate behavior rather than a failure of $\eta_k$.
Suppressing the remainder with $\eta=0$, by contrast, causes marked degradation.
Together, these results show that the remainder should be suppressed only when its subspace is poorly observed, not whenever the motion violates the wheel model.

\begin{table}[!t]
\centering
\caption{Ablation results on stress sequences.}
\label{tbl:two_ablation_metrics}
\scriptsize
\setlength{\tabcolsep}{2.1pt}
\renewcommand{\arraystretch}{1.0}
\begin{threeparttable}
\begin{tabular}{@{}llcccccc@{}}
\toprule
\textbf{Scenario} & \textbf{Metric} & \textbf{TWO} & \textbf{$\eta=1$} & \textbf{$\gamma=1$} & \textbf{$\eta=0$} & \textbf{$\beta=0$} & \textbf{w/o fb}\\
\midrule
\multirow{2}{*}{Slope} & ATE & \underline{0.025} & \textbf{0.024} & 0.033 & 0.099 & 0.053 & 0.027\\
& RPE & \underline{0.017} & \textbf{0.015} & 0.019 & 0.049 & 0.022 & 0.019\\
\midrule
\multirow{2}{*}{\begin{tabular}[c]{@{}l@{}}Forward Slip\\+ Weak Geometry\end{tabular}} & ATE & \textbf{0.043} & 0.051 & \underline{0.047} & 0.048 & 0.361 & 0.069\\
& RPE & \underline{0.018} & 0.023 & 0.019 & \textbf{0.017} & 0.058 & \textbf{0.017}\\
\midrule
\multirow{2}{*}{Side Slip} & ATE & 0.072 & \textbf{0.067} & \underline{0.070} & 0.164 & 0.176 & 0.076\\
& RPE & \underline{0.032} & \textbf{0.031} & \textbf{0.031} & 0.059 & 0.058 & 0.035\\
\midrule
\multirow{2}{*}{\begin{tabular}[c]{@{}l@{}}Lift/Pose\\Perturbation\end{tabular}} & ATE & \underline{0.042} & \underline{0.042} & \textbf{0.041} & 0.115 & 0.284 & 0.063\\
& RPE & \textbf{0.050} & 0.054 & \underline{0.051} & 0.143 & 0.180 & 0.077\\
\midrule
\multirow{2}{*}{Average} & ATE & \textbf{0.0455} & \underline{0.0460} & 0.0478 & 0.1065 & 0.2185 & 0.0588\\
& RPE & \textbf{0.0293} & 0.0307 & \underline{0.0300} & 0.0670 & 0.0795 & 0.0370\\
\bottomrule
\end{tabular}
\begin{tablenotes}[flushleft]\scriptsize
\item \textit{Note:} ATE/RPE are in meters, and lower values are better. Best results are in bold and second-best results are underlined.
Average rows are arithmetic means over the four stress sequences for each metric separately.
\end{tablenotes}
\end{threeparttable}
\end{table}

\textbf{Degeneracy-aware damping (C2).}
The role of $\gamma_k$ is complementary: it controls how strongly the correction updates the inflation magnitude.
Across the four sequences, fixing $\gamma=1$ changes average ATE/RPE from $0.046/0.029$\,m to $0.048/0.030$\,m.
The largest degradation occurs on Slope, while the two methods remain close on the other sequences.
In Forward Slip + Weak Geometry, Fig.~\ref{fig:ablation_selective_behavior} further shows a larger attitude excursion when damping is disabled.
Thus, $\gamma_k$ provides a mild geometry-dependent safeguard without imposing a conservative scale on all events.
C1 and C2 therefore separate $\eta_k$-based direction selection from $\gamma_k$-based magnitude adjustment.

\textbf{Intra-frame feedback (C3).}
The final hypothesis concerns when the inferred covariance is applied.
Without intra-frame feedback, it is used only after the current update.
This retains the one-frame delay of standard covariance adaptation and leaves the current-frame prior overconfident during bursty motions.
Omitting feedback increases ATE by $1.60\times$ in Forward Slip + Weak Geometry and $1.50\times$ in Lift/Pose Perturbation.
This supports the need for the same-frame covariance repair loop.

To characterize the accuracy-latency tradeoff of the intra-frame loop, Table~\ref{tbl:iteration_sensitivity} reports feedback cost and parameter sensitivity.
Avg.\ iter.\ is the mean additional repair passes per frame.
It stays below one since most frames satisfy the stopping criterion after the initial update, with extra repairs concentrated around kinematic violations.
One feedback iteration already captures most of the accuracy gain; the default limit of two averages $0.49$ passes and $12.575/32.902$\,ms mean/P95.

\begin{table}[!t]
\centering
\begingroup
\caption{Feedback-iteration sensitivity and embedded runtime.}
\label{tbl:iteration_sensitivity}
\scriptsize
\setlength{\tabcolsep}{4pt}
\renewcommand{\arraystretch}{1.08}
\begin{threeparttable}
\begin{tabular*}{\columnwidth}{@{\extracolsep{\fill}}lcccc@{}}
\toprule
\textbf{Max. iter.} & \textbf{ATE RMSE} & \textbf{Avg. iter.} & \textbf{Mean [ms]} & \textbf{P95 [ms]}\\
\midrule
$0$ & 0.055 & 0.00 & 6.665 & 15.547\\
$1$ & 0.045 & 0.29 & 11.096 & 26.344\\
$2$ & 0.046 & 0.49 & 12.575 & 32.902\\
\bottomrule
\end{tabular*}
\begin{tablenotes}[flushleft]\scriptsize
\item Accuracy entries are four-sequence ATE RMSE means in meters.
Feedback iterations exclude the initial update; times are total per-frame update times on Raspberry Pi~4B.
\end{tablenotes}
\end{threeparttable}
\endgroup
\vspace{-1.8ex}
\end{table}

To separate directional adaptation from fixed process-noise tuning, Table~\ref{tbl:q0_sensitivity} compares full TWO against a static control that retains $\mathbf Q_w+\mathbf Q_0$ but disables $\mathbf Q_{\mathrm{infl}}$, across $Q_0\in\{0.1,0.5,1,2,10\}\times$ on the four stress sequences.
At every scale, TWO reduces stress-sequence ATE by 74.3--83.2\%; the gain therefore cannot be reproduced by globally inflating baseline process uncertainty.

\begin{table}[!t]
\centering
\begingroup
\caption{$\mathbf Q_0$ sensitivity: static vs. full TWO on the stress sequences.}
\label{tbl:q0_sensitivity}
\scriptsize
\setlength{\tabcolsep}{4pt}
\renewcommand{\arraystretch}{1.08}
\begin{threeparttable}
\begin{tabular*}{\columnwidth}{@{\extracolsep{\fill}}lccccc@{}}
\toprule
\textbf{$\mathbf{Q}_0$ scale} & \textbf{$0.1\times$} & \textbf{$0.5\times$} & \textbf{$1\times$} & \textbf{$2\times$} & \textbf{$10\times$}\\
\midrule
Static ATE (m) & 0.224 & 0.212 & 0.219 & 0.211 & 0.226\\
Full ATE (m) & 0.058 & 0.047 & 0.046 & 0.042 & 0.038\\
Reduction & 74.3\% & 77.7\% & 79.0\% & 80.1\% & 83.2\%\\
\bottomrule
\end{tabular*}
\begin{tablenotes}[flushleft]\scriptsize
\item ATE RMSE means over the four stress sequences, in meters.
Static retains $\mathbf Q_{w,i}+\mathbf Q_0$ with $\mathbf Q_{\mathrm{infl},i}$ disabled, and Full is TWO with $\mathbf Q_{\mathrm{infl},i}$ enabled. Reduction is the relative ATE decrease from Static to Full.
\end{tablenotes}
\end{threeparttable}
\endgroup
\vspace{-1.8ex}
\end{table}

\section{Conclusion}

In this paper, we presented ToF--Wheel Odometry (TWO), a tightly-coupled filtering framework that improves robustness for ground robots under limited sensing capabilities.
By integrating wheel kinematics with ToF observations, TWO addresses the interplay between transient motion violations and geometric degeneracy.
The core of our approach lies in a degeneracy-aware, directional covariance-shaping strategy that modulates process noise based on geometric observability.
The novel intra-frame feedback mechanism allows TWO to recover rapidly from bursty motion disturbances and improve robustness in challenging conditions.

Compared to fixed and scalar-$\mathbf Q$ ToF--wheel baselines, TWO achieves superior accuracy and robustness, particularly during extreme motion events, and the gain comes from structured covariance adaptation rather than sensor fusion alone.
The ablations further show that this benefit does not come from a single conservative tuning choice. When the off-manifold motion is observable, the adaptive gates remain open and admit it. When the correction is ambiguous under degraded ToF geometry, however, TWO suppresses the update and avoids attitude contamination.
This selectivity is the central practical advantage of TWO: it can switch between admitting real kinematic violations and rejecting geometrically unsupported corrections without requiring sequence-specific parameter retuning.

\bibliographystyle{IEEEtran}
\bibliography{TWO}

\end{document}